%% file: bim.tex
\documentclass[11pt]{article}

\usepackage{colacl}
\usepackage{graphicx,psfrag}
\usepackage{amsthm,amsmath}

\newcommand*\path[1]{\mathit{Path}_{#1}}

\newcommand*\SimultMatch{\textit{SimultMatch}}
\newcommand*\leftar[1]{\overset{\rightarrow}{#1}}
\newcommand*\rightar[1]{\overset{\leftarrow}{#1}}

\newcommand*\leftfsa{\overset{\rightarrow}{A}}
\newcommand*\rightfsa{\overset{\leftarrow}{A}}
\newcommand*\Cpp{C\raisebox{.25ex}{\footnotesize++}}

\newenvironment{abstractquote}{\begin{list}{}{\setlength{\rightmargin}{0.25in}\setlength{\leftmargin}{0.25in}}\item[]\small}{\end{list}}

\title{A Bimachine Compiler for Ranked Tagging Rules}

\author{Wojciech Skut, Stefan Ulrich and Kathrine Hammervold\\
Rhetorical Systems\\ 4 Crichton's Close, Edinburgh\\ EH8 8DT\\ Scotland\\
{\tt wojciech@rhetorical.com}}

\begin{document}
\maketitle
\bibliographystyle{acl}

\begin{abstract}
\begin{abstractquote}

This paper describes a novel method of compiling ranked tagging rules
into a deterministic finite-state device called a {\em bimachine}. The
rules are formulated in the framework of {\em regular rewrite
operations} and allow unrestricted regular expressions in both left
and right rule contexts. The compiler is illustrated by an application
within a speech synthesis system.
\end{abstractquote}
\end{abstract}

\input{motivation}

\input{formalization}

\input{extensions}

\input{applications}

\input{conclusions}

\bibliography{../references/references}

\end{document}

%% file: motivation.tex
\section{Motivation}\label{sec:motivation}

In rule-based tagging, linguistic objects (e.g. phonemes, syllables,
or words) are assigned linguistically meaningful labels based on the
context. Each instance of label assignment is licensed by a {\em
tagging rule} typically specifying that label $\psi$ can be
assigned to item $\phi$ if $\phi$ is preceded by a pattern $\lambda$
and followed by a pattern $\rho$.  The patterns $\lambda$ and $\rho$
are usually formulated as regular expressions over the input alphabet,
but may also range over output labels.

The nature of the tagging task suggests a formalisation in terms of
\emph{finite-state transducers} (FSTs). More precisely, the task can
be viewed as an instance of string rewriting. In this framework, a
tagging rule is interpreted as a {\em regular rewrite rule}
$\phi \rightarrow \psi / \lambda\mathop{\_}\rho$.

Several methods have been proposed for the compilation of such rules
into FSTs
\cite{Kaplan:Kay:94,Mohri:Sproat:96,Gerdemann:VanNoord:99}.
A rewrite rule is converted into a number of transducers, which are
combined by means of transducer composition, yielding an FST that
implements the actual rewrite operation.

Typically, tagging is carried out by a set of rules $R_i:\phi_i
\rightarrow \psi_i / \lambda_i\mathop{\_}\rho_i$, $i=1\ldots n$, which may 
overlap and/or conflict. A regular rule compiler should not just
convert the rules into separate transducers $T_1\ldots T_n$. For
efficiency reasons, it is highly desirable to convert them into a
single machine in a way that determines how rule conflicts should be
resolved.

There are two basic options as to how the rule transducers can be
combined.
\begin{itemize}
\item The rules can be associated with numerical {\em costs}, which 
translate as transition weights in the compilation step.  In this
formalisation, the union $T_1 \cup \ldots \cup T_n$ is a {\em weighted
finite-state transducer} (WFST). This transducer is typically
non-deterministic, but the weights make it possible to find the
optimal path efficiently (as an instance of the single-source
shortest paths problem).

\item The rules can be explicitly ranked. In such a case, 
{\em priority union}  \cite{Karttunen:98}, or an
equivalent operator, can be used to combine $T_1,\ldots,T_n$ into a
single unambiguous FST which is then turned into a deterministic
device
\cite{Skut:ea:04}.
\end{itemize}
The work reported here pursues the latter strategy. Although less
powerful and flexible than e.g. probabilistic approaches, it has
the advantage of efficiency: once the rules have been compiled,
rewriting an input sequence of $t$ symbols boils down to $t$ lookups
in a transition table ($2t$ in case of a bimachine, see below).
\footnote{With hand-written rules, the
simple ranking is actually an advantage since a more complex
rule interaction typically affects the transparency of the system.}

Several compilation methods have been proposed for creating a
deterministic machine out of a set of rules
\cite{Laporte:97,Roche:Schabes:95,Hetherington:01}. However, most of
them impose strong restrictions on the form of contextual constraints:
$\lambda$ and $\rho$ are restricted to single symbols
\cite{Hetherington:01}, or acyclic regular expressions
\cite{Laporte:97}. 

Skut et al.~\shortcite{Skut:ea:04} describe a more powerful rewrite
rule compiler that does not impose such constraints on $\phi$,
$\lambda$ and $\rho$. Each rule $R_i$ is compiled into an unambiguous
FST $T_i$ that inserts a marker $\#_i$ at the beginning of every match
of $\phi_i$ preceded by an instance of $\lambda_i$ and followed by an
instance of $\rho_i$. While $\rho_i$ and $\lambda_i$ may contain
markers inserted by other rules, $\phi_i$ must be marker-free. The
composition $T_1 \circ \ldots \circ T_n$ of the rule transducers
yields an FST that inserts rule markers into the input string,
resolving rule conflicts according to the explicit ranking of
rules. Composed with a transducer that performs the actual rewrite
operation, it produces an unambiguous FST which implements the
required combination of rules.

Two problems arise with this approach. 
\begin{itemize} 
\item Although the resulting FST is unambiguous (i.e., implements 
a function), it may be non-determinisable
\cite{Mohri:97,Laporte:97}.

\item The composition operation used to combine the ranked rule 
transducers quickly creates large non-deterministic FSTs, resulting in
slow compilation  and high memory requirements.
\end{itemize}
The remedy to the first problem is straightforward: since the
resulting FST implements a function, it can be compiled into a
bimachine, i.e. an aggregate of a left-to-right and a right-to-left
deterministic finite-state automaton (FSA) associated with an output
function. The application of such a bimachine to a string involves
running both automata (in the respective directions) and determining
the symbols emitted by the output function (cf. 
section~\ref{sec:defs}).

The simplest option is thus first to create the rule transducers, then
compose them into a (non-deterministic) FST, and finally apply a
bimachine construction method \shortcite{Roche:Schabes:96}. However,
such a solution will not eliminate the inefficiency caused by
expensive rule composition.

Thus, we have developed a compilation method that constructs the
left-to-right and the right-to-left automaton of the resulting
bimachine directly from the patterns without having to construct and
then to compose the rule transducers. The efficiency of the compiler
is increased by employing \emph{finite-state automata} instead of
FSTs, since algorithms used to process FSAs are typically faster than
the corresponding transducer algorithms. Furthermore, the resulting
(intermediate) structures are significantly smaller than in the case
of FSTs.  This leads to much faster compilation and smaller
finite-state machines.

%% file: formalization.tex
\section{Formalization}

\subsection{Definitions and Notation}
\label{sec:defs}

In the following definitions, $\Sigma$ denotes a finite input
alphabet. $\Delta$ is a finite output alphabet.

A \emph{deterministic finite-state automaton} (DFSA) is a quintuple
$A=(\Sigma, Q, q_0,\delta, F)$ such that:
\begin{description}
\item[$Q$] is a finite set of states;
\item[$q_0 \in Q$] is the initial state of $A$;
\item[$\delta: Q \times \Sigma \rightarrow Q$] is the transition function
of $A$; 
\item[$F \subset Q$] is a non-empty set of final states.
\end{description}

A {\em sequential transducer} (ST) is defined as a 7-tuple $T=(\Sigma,
\Delta, Q, q_0, \delta, \sigma, F)$ such that $(\Sigma, Q, q_0,\delta, F)$ 
is a DFSA, and $\sigma(q, a)$ is the output associated with the
transition from state $q$ via symbol $a$ to state $\delta(q,a)$.

The functions $\delta$ and $\sigma$ can be extended to the domain $Q
\times \Sigma^*$ by the recursive definition: $\delta^*(q,\epsilon)=q$, 
$\delta^*(q,wa)=\delta(\delta^*(q, w), a)$, $w\in\Sigma^*$, $a \in \Sigma$ and
$\sigma^*(q,\epsilon) = \epsilon$, $\sigma^*(q, wa) = \sigma^*(q,
w)\sigma(\delta^*(q,w),a)$.

A {\em bimachine} $B$ is a triple $(\leftfsa, \rightfsa, h)$ such that:
\begin{description}
\item[$\leftfsa = (\Sigma, \leftar{Q}, \leftar{q}_0, \leftar{\delta})$]
is a left-to-right FSA (there is no concept of final states in a bimachine);

\item[$\rightfsa = (\Sigma, \rightar{Q}, \rightar{q}_0, \rightar{\delta})$]
is a right-to-left FSA;
\item[$h: \leftar{Q} \times \Sigma \times \rightar{Q} \rightarrow \Delta^*$]
is the {\em output function}. 
\end{description}
Applied to a string $u=a_1\ldots a_t$, $B$ produces a string 
$v=b_1\cdot \ldots \cdot b_t$, such that $b_i \in \Delta^*$ is defined
as follows:
\[
b_i = h(\leftar{\delta}(\leftar{q}_0, a_1 \dots a_{i-1}),a_i,\rightar{\delta}(\rightar{q}_0, a_{t}\dots a_{i+1}))
\]
An unambiguous finite-state transducer can always be converted into a
bimachine
\cite{Berstel:79,Roche:Schabes:96}. This property makes bimachines
an attractive tool for deterministic processing, especially since 
not all unambiguous transducers are determinisable (sequentiable).

\subsection{Rules, Rule Ordering and Priorities}
\label{sec:formalism}

As mentioned in section~\ref{sec:motivation}, the tagging rule
formalism is formulated in the framework of regular rewrite rules
\cite{Kaplan:Kay:94}. Input to the rule compiler consists of a set of
rules.
\begin{eqnarray*}
R_1 :  \phi_1 & \rightarrow & \psi_1 / \lambda_1\mathop{\_}\rho_1\\
		      & \vdots & \\
R_n :  \phi_n & \rightarrow & \psi_n / \lambda_n\mathop{\_}\rho_n
\end{eqnarray*}
A rule $\phi \rightarrow \psi / \lambda\mathop{\_}\rho$ states that
label $\psi$ is assigned to object $\phi$ (called the {\em focus} of
the rule) if $\phi$ is preceded by a left context $\lambda$ and
followed by a right context $\rho$. The context descriptions
$\lambda$, $\rho$ and $\phi$ are formulated as regular expressions
over the input alphabet $\Sigma$. 

The rules may conflict, in which case the ambiguity is resolved based
on the order of the rules in the grammar. If a rule $R_i$ fires for an
object $s_k$ in the input sequence $s_1\ldots s_t$, it blocks the
application of all rules $R_j$, $j > i$, to $s_k$.

Thus the operational semantics of the rules may be stated as
follows: A rule $R_i$ fires if:
\begin{itemize}

\item[(a)] no other rule $R_j$, $j < i$, is applicable in the same context;

\item[(b)] the substring $s_1\ldots s_{k-1}$ matches the regular expression $\Sigma^*\lambda_i$;

\item[(c)] the substring $s_k\ldots s_t$ matches the regular expression $\phi_i\rho_i \Sigma^*$.

\end{itemize}
This basic formalism imposes two conditions on the rules:
\begin{itemize}

\item $\lambda$ and $\rho$ are regular expressions
	over $\Sigma$;

\item $\phi \in \Sigma$ is a single object.

\end{itemize}
These two assumptions restrict the expressive power of the formalism
compared to general regular rewrite
rules~\cite{Kaplan:Kay:94,Mohri:Sproat:96} in that they
do not allow output symbols on either side of the context and only
admit rule foci of length one. Although these restrictions are
essential for the initial basic formalism, we show in
section~\ref{sec:extensions} how to extend it so that the compiler can
accept rules with both input and output symbols in the left
context. As for the length of the focus, it is important to bear in
mind that the formalism is primarily intended for tagging rules which
usually do not cover longer foci.

\subsection{Matching of Context Patterns}
\label{sec:simultmatch}

The basic idea in the new compilation method is to convert the
patterns $\lambda_i$ and $\rho_i \phi_i$, $i=1,\ldots n$, directly
into the left-to-right and right-to-left acceptor of a bimachine
without having to perform the fairly expensive operations required by
the transducer-based approaches.

Key to the solution is the function
$\SimultMatch_{\beta_1\ldots\beta_n}: \Sigma^t \rightarrow
(2^{\text{\textbf N}})^{t+1}$, $t \in \textbf{N}$, which, given a collection
$\{\beta_1,\ldots,\beta_n\}$ of regular expressions, maps a sequence of
symbols $s_1\ldots s_t$ to a sequence of $t+1$ sets of indices
corresponding to the matching patterns at the respective position
(position $0$ corresponds to the beginning of the string, $I$ denotes
the set $\{1,\ldots,n\}$ of rule indices):
\begin{multline*}
\SimultMatch_{\beta_1\ldots\beta_n} (s_1\ldots s_t)[k] =\\
\{j\in I: \Sigma^*\beta_j \mbox{ matches } s_1\ldots s_{k}\}
\end{multline*}
This
construct can be implemented as a pair $(A_{\beta_1\ldots\beta_n},
\tau)$ such that:
\begin{description}
\item[$A_{\beta_1\ldots\beta_n}= (\Sigma, Q, q_0, \delta, F)$] is a
finite-state automaton that encodes in its states ($Q$) information
about the matching patterns.

\item[$\tau: Q \rightarrow 2^I$] is a function mapping the states of
$A$ to sets of indices corresponding to matching regular expressions:
$\tau(\delta^*(q_0, w)) = \{ j \in I: \Sigma^*\beta_j\mbox{ matches }w\}$.

\end{description}
In order to construct $\SimultMatch_{\beta_1\ldots\beta_n}$, we
introduce a marker symbol $\$_j \not\in \Sigma$ for each $\beta_j$.
Let $\Sigma_\$ = \Sigma \cup \bigcup_{j\in I} \{\$_j\}$ be the
extended alphabet. Let $\tilde{A}=(\Sigma_\$, \tilde{Q}, \tilde{q_0}, \tilde{\delta}, \tilde{F})$ be a
deterministic finite acceptor for the regular expressions
$ \Sigma^*\bigcup_{j\in I} \beta_j \$_j$.

An important property of the automaton $\tilde{A}$ is that $w\in \Sigma^*$ is an
instance of a pattern $\Sigma^* \beta_j$ if and only if $q =
\tilde{\delta}^*(\tilde{q_0}, w)$ is defined and there exists a transition from $q$ by
$\$_j$ to a final state: $\tilde{\delta}(\tilde{\delta}^*(\tilde{q_0}, w), \$_j) \in F$.
Now we can define the function $\tau: Q \rightarrow 2^I$:
\[\tau(q) = \{j \in I: (q, \$_j) \in \mathit{Dom}(\tilde{\delta}) \wedge\tilde{\delta}(q, \$_j) \in F \} \]

Obviously, if $\tilde{A}$ enters state $q$ after consuming a string $w\in
\Sigma^*$, $\tau(q)$ is the set of all indices $j$ such that $w$
matches $\Sigma^*\beta_j$.

The automaton $A_{\beta_1\ldots\beta_n} = (\Sigma, Q, q_0, \delta, Q)$
can now be constructed from $\tilde{A}$ by restricting it to the alphabet
$\Sigma$ (which includes trimming away the unreachable states)
and making all its states final so that it accepts all strings $w \in
\Sigma^*$:
\begin{itemize}
\item[] $Q=\{ q \in \tilde{Q}: \exists w \in \Sigma^*\quad \tilde{\delta}^*(\tilde{q_0}, w) = q\}$
\item[] $q_0=\tilde{q_0}$ 
\item[] $\delta = \tilde{\delta} | _{Q \times \Sigma \times Q}$.
\end{itemize}
The resulting construct $(A_{\beta_1\ldots\beta_n}, \tau)$ makes it
possible to simultaneously match a collection of regular expressions.

\subsection{Bimachine Compilation}
\label{sec:compiler}

Using the construct $\SimultMatch$, we can determine all the matching
left and right contexts at any position $k$ in a string $w=a_1\ldots
a_t$. The value of $\SimultMatch_{\lambda_1\ldots\lambda_n}(w)[k-1]$
is the set of all rule indices $i$ such that $\lambda_i$ matches the
string $a_1\ldots
a_{k-1}$. $\SimultMatch_{(\phi_1\rho_1)^{-1}\ldots(\phi_n\rho_n)^{-1}}(w^{-1})[t-k]$
is the set of all rule indices $i$ such that $\phi_i\rho_i$ matches
the remainder $s_k\ldots s_t$ of $w$. Obviously, the intersection
$\SimultMatch_{\lambda_1\ldots\lambda_n}(w)[k-1] \cap
\SimultMatch_{(\phi_1\rho_1)^{-1}\ldots(\phi_n\rho_n)^{-1}}(w^{-1})[t-k]$
is exactly the set of all matching rules at position $k$. The
minimal element of this set is the index of the firing rule.%
\footnote{
In order to ensure that the above formula is always valid, we assume
that the rule with the highest index ($R_n$) matches all left and
right contexts (i.e., $\lambda_n=\rho_n=\Sigma^*, \phi_n =
\bigcup_{\sigma \in \Sigma}
\{\sigma\}$), and $\psi_n$ is a vacuous action. If
none of the other rules fire, the formalism defaults to $R_n$.  }

Now if $(\leftfsa, \leftar{\tau}):=
\SimultMatch_{\lambda_1\ldots\lambda_n}$, and
$(\rightfsa,\rightar{\tau}) :=
\SimultMatch_{(\phi_1\rho_1)^{-1}\ldots(\phi_n\rho_n)^{-1}}$, then
the tagging task is performed by the bimachine
$B=(\leftfsa,\rightfsa,h)$, where the output function $h:
\leftar{Q}\times \Sigma \times \rightar{Q} \rightarrow \Delta^*$ is
defined as follows:
\begin{eqnarray*}
h(\leftar{q}, a, \rightar{q})&=&\psi_{\min(\leftar{\tau}(\leftar{q}) \; \cap \; \rightar{\tau}(\rightar{\delta}(\rightar{q},a)))}
\end{eqnarray*}
The output function can be either precompiled (e.g., into a hash
table), or -- if the resulting table is too large -- the intersection
operation can be performed at runtime, e.g. using a bitset encoding of
sets.\footnote{ In the actual implementation of the tagger, $h$ has
been replaced by a function $g: \leftar{Q} \times \rightar{Q}
\rightarrow \Delta^*$ defined as:
\begin{eqnarray*}
g(\leftar{q}, \rightar{q})&=&\psi_{\min(\leftar{\tau}(\leftar{q}) \; \cap \; \rightar{\tau}( \rightar{q}))}
\end{eqnarray*}
The translation of the $k$-th symbol in a string $w=a_1\ldots a_t$ is
then determined by the formula 
\[g(\leftar{\delta}(\leftar{q}_0,
a_1\ldots a_{k-1}),\rightar{\delta}(\rightar{q}_0, a_t\ldots a_{k}))\]
which is easier to compute.}

%% file: extensions.tex
\section{Extensions}
\label{sec:extensions}

The compiler introduced in the previous section can be extended
to handle more sophisticated rules and search\slash control strategies.

\subsection{Output Symbols in Left Contexts}
\label{sec:leftoutputs}

The rule formalism can be extended by including output symbols in the
left context of a rule. This extra bit is added in the form of a
regular expression $\pi$ ranging over the output symbols, which can be
represented by rule IDs $r_k\in I$. The rules then look as follows:
\begin{eqnarray*}
R_i = \phi_i  & \rightarrow & \psi_i / \pi_i:\lambda_i\mathop{\_}\rho_i
\end{eqnarray*}
Such a rule fires at a position $k$ in string $s_1\ldots s_t$ if an
extra condition (d) holds in addition to the conditions (a)--(c),
formulated in section~\ref{sec:formalism}:
\begin{itemize}
\item[(d)] The IDs $r_1\ldots r_{k-1}$ of the firing rules match 
$I^*\pi_i$.
\end{itemize}
In order to enforce condition (d), we use the $\SimultMatch$ construct
introduced in section~\ref{sec:simultmatch}. For that, the patterns
$\pi = \{\pi_1,\ldots, \pi_n\}$ are compiled into an instance of
$\SimultMatch_{\pi} = (A_\pi, \tau_\pi)$.  $A_\pi$ is an FSA, so
$A_\pi = (I, Q_\pi, q^{\pi}_0, \delta_\pi)$. It follows from the
construction of $\SimultMatch_{\pi}$ that the function $\tau_\pi:
Q_\pi\rightarrow 2^I$ has the following property:
\begin{multline*}
\tau_\pi(\delta_\pi^*(q_0^\pi, r_1\ldots r_k)) =\\
\{ j \in I: r_1\ldots r_{k} \mbox{ matches } \pi_j \}
\end{multline*}
In other words, an action $\psi_{r_k}$ is admissible at position $k$
if $r_k \in \tau_\pi(\delta_\pi^*(q_0, r_1\ldots\allowbreak
r_{k-1}))$.  Thus, the tagging task (according to the extended
strategy (a)--(d)) is performed by the formal machine
$M=(\leftfsa,A_{\pi},\rightfsa,h)$, where $\leftfsa$ and $\rightfsa$
are as in section~\ref{sec:compiler}, and $h: \leftar{Q} \times
Q_{\pi} \times \Sigma \times \rightar{Q}\rightarrow \Delta^*$ is
defined as follows:%
\footnote{
In order to make sure the definition of $h$ is always valid, we assume
that the rule with the highest index ($n$) matches all possible
contexts (i.e., $\pi_n = I^*, \lambda_n = \rho_n = \Sigma^*$ and
$\phi_n =
\bigcup_{\sigma \in \Sigma} \{\sigma\}$). 
}
\[h(q,q^{\pi},a,q') = \psi_{r_k},\]
where 
\[r_k  :=  \min(\leftar{\tau}(q) \cap \tau_{\pi}(q^{\pi}) 
\cap \rightar{\tau}(\rightar{\delta}(q',a)) )\]
In this formula, $\leftar{q}$ and $\rightar{q}$ are as in the basic
bimachine introduced in section~\ref{sec:compiler}. $q^{\pi}$ 
is the state of $A_\pi$ after consuming the rule IDs $r_1\ldots r_{k-1}$:
$q^\pi := \delta_\pi^*(q_0^\pi, r_1\ldots r_{k-1})$.

In order to determine the tagging actions for an input sequence
$w=a_1\ldots a_t$, the automaton $\rightfsa$ is first run on $w^{-1}$.
Then both $\leftfsa$ and $A_\pi$ are run on $w$ in parallel. In each
step $k$, the states $\leftar{\delta}(\leftar{q}_0, a_1\ldots
a_{k-1})$, $\rightar{\delta}(\rightar{q}_0, a_t\ldots a_{k})$ as well
as the sequence $r_1\ldots r_{k-1}$ of already executed actions are
known, so that the $\psi_{r_k}$'s can be determined incrementally from
left to right.

\subsection{Alternative Control Strategies}
\label{sec:altercontrol}

Our rule compilation method is very flexible with respect to control
strategies. By intersecting the sets of rule IDs
$\leftar{\tau}(\leftar{\delta}(\leftar{q}_0, a_1\ldots a_{k-1}))$ and
$\rightar{\tau}(\rightar{\delta}(\rightar{q}_0,\allowbreak a_t\ldots
a_k))$, $1
\leq k \leq t$, one can determine the set of all matching rules for each
position in the input string. In the formalism presented in
section~\ref{sec:compiler}, only one rule is selected, namely the one
with the minimal ID. This is probably the most common way of handling
rule conflicts, but the formalism does not exclude other control
strategies.

\begin{description}
\item[Simultaneous matching of all rules:] This strategy is particularly
useful in the machine-learning scenario, e.g. in computing scores in
transformation-based learning \cite{Brill:95}. Note that the simple
context rules used by Brill~\shortcite{Brill:95} may be mixed with
more sophisticated hand-written heuristics formulated as regular
expressions while still being subject to scoring. As shown in
section~\ref{sec:formalism}, taggers using unrestricted regular
context constraints are not sequentiable, and thus cannot be
implemented using ST-based rule compilation methods
\cite{Roche:Schabes:95}.

\item[N-best/Viterbi search:] Instead of a strict ranking, the 
rules may be associated with probabilities or scores such that the
best sequence of actions is picked based on global, per-sequence,
optimisation rather than on a sequence of greedy local decisions. In
order to implement this, we can use the extended formalism introduced
in section~\ref{sec:leftoutputs} with a slight modification: in each
step, we keep $N$ best-scoring paths rather than just the one
determined by the selection of the locally optimal action $\psi_{r_k}$
for $1 \leq k \leq t$.

\end{description}

%% file: applications.tex
\section{An Application}\label{sec:applications}

In this section, we describe how our bimachine compiler has been
applied to the task of homograph disambiguation in the rVoice speech
synthesis system.

Each module in the system adds information to a structured relation
graph (HRG), which represents the input sentence or \emph{utterance} to
be spoken \cite{Taylor:Black:Caley:01}. The HRG consists of several
\emph{relations}, which are structures such as lists or trees over a set
of \emph{items}. The homograph tagger works on a \emph{list} relation,
where the items represent words. Each item has a feature structure
associated with it, the most relevant features for our application
being the \emph{name} feature representing the normalised word and the
\emph{pos} feature representing the part-of-speech (POS) of the word.

The assignment of POS tags is done by a statistical tagger (a trigram
HMM). Its output is often sufficient to disambiguate homographs, but
in some cases POS cannot discriminate between two different
pronunciations, as in the case of the word {\em lead}: {\em they took
a 1-0 lead} vs. {\em a lead pipe} (both nouns). Furthermore, the
statistical tagger turns out to be less reliable in certain
contexts. The rule-based homograph tagger is a convenient way of
fixing such problems.

The grammar of the homograph tagger consists of a set of ordered rules
that define a mapping from an item to a \emph{sense ID}, which
uniquely identifies the phonetic transcription of the item in a
pronunciation lexicon.

For better readability, we have changed the rule syntax. Instead of
$\phi\rightarrow \psi / \lambda\mathop{\_}\rho$, we write:
\begin{eqnarray*}
\lambda / \phi / \rho \rightarrow \psi
\end{eqnarray*}
where $\lambda$, $\phi$ and $\rho$ are regular expressions over
feature structures. The feature structures are written $[f_1=v_1\ldots
f_k=v_k]$, where $f_i$ is a feature name and $v_i$ an atomic value or
a disjunction of atomic values for that feature. Each attribute-value
pair constitutes a separate input alphabet symbol. The alphabet also
contains a special default symbol that denotes feature-value pairs not
appearing in the rules. The symbol $\psi$ stands for the action of
setting the sense feature of the item to a particular sense ID.

The following are examples of some of the rules
that disambiguate between the different senses of \emph{suspects}
({\tt sense=1} is the noun reading, {\tt sense=2} the verb reading):
{\small
\begin{verbatim}
[name=that] 
    / [name=suspects] / 
                       -> [sense=2];

([pos=dt|cd]|[name=terror]) 
         / [name=suspects]/ 
                       -> [sense=1];
/ [name=suspects] / 
              [name=that] 
                       -> [sense=2];
/ [name=suspects] /    -> [sense=1];
\end{verbatim}
}
Note that the last rule is a default one that sets \emph{sense}
to 1 for all instances of the word \emph{suspects} where none of
the other rules fire. 

To explain the interaction of the rules, we will look at the 
following example:

\begin{center}
\emph{the}$_1$ \emph{terror}$_2$ \emph{suspects}$_3$ \emph{that}$_4$  \emph{were}$_5$  \emph{in}$_6$ \emph{court}$_7$\\
\end{center}
We can see that the second and the third rule match the context of
word~3. The rule associated with the lower index fires, resulting in
the value of \emph{sense} being set to 1 on the item.

\begin{figure}
\newcommand{\ftextsize}{\fontsize{8}{8}\selectfont}
\newcommand{\flabelsize}{\fontsize{8}{8}\selectfont}
\psfrag{35000}[Tr][Tr]{{\ftextsize \textit{Tr}\ \ 35000}}
\psfrag{30000}[Tr][Tr]{{\ftextsize 30000}}
\psfrag{25000}[Tr][Tr]{{\ftextsize 25000}}
\psfrag{20000}[Tr][Tr]{{\ftextsize 20000}}
\psfrag{15000}[Tr][Tr]{{\ftextsize 15000}}
\psfrag{10000}[Tr][Tr]{{\ftextsize 10000}}
\psfrag{5000}[Tr][Tr]{{\ftextsize 5000}}
\psfrag{1000}[Tr][Tr]{{\ftextsize \textit{S}\ \ 1000}}
\psfrag{900}[Tr][Tr]{{\ftextsize 900}}
\psfrag{800}[Tr][Tr]{{\ftextsize 800}}
\psfrag{700}[Tr][Tr]{{\ftextsize 700}}
\psfrag{600}[Tr][Tr]{{\ftextsize 600}}
\psfrag{500}[Tr][Tr]{{\ftextsize 500}}
\psfrag{400}[Tr][Tr]{{\ftextsize 400}}
\psfrag{300}[Tr][Tr]{{\ftextsize 300}}
\psfrag{200}[Tr][Tr]{{\ftextsize 200}}
\psfrag{100}[Tr][Tr]{{\ftextsize 100}}
\psfrag{120}[Tr][Tr]{{\ftextsize \textit{t}\ \ 120}}
\psfrag{100}[Tr][Tr]{{\ftextsize 100}}
\psfrag{80}[Tr][Tr]{{\ftextsize 80}}
\psfrag{60}[Tr][Tr]{{\ftextsize 60}}
\psfrag{35}[Tr][Tr]{{\ftextsize 35}}
\psfrag{40}[Tr][Tr]{{\ftextsize 40}}
\psfrag{40.0001}[Tl][Tl]{{\ftextsize 40\ \ \textit{n}}}
\psfrag{35}[Tr][Tr]{{\ftextsize 35}}
\psfrag{30}[Tr][Tr]{{\ftextsize 30}}
\psfrag{25}[Tr][Tr]{{\ftextsize 25}}
\psfrag{20}[Tr][Tr]{{\ftextsize 20}}
\psfrag{15}[Tr][Tr]{{\ftextsize 15}}
\psfrag{10}[Tr][Tr]{{\ftextsize 10}}
\psfrag{5}[Tr][Tr]{{\ftextsize 5}}
\psfrag{0}[Tr][Tr]{{\ftextsize 0}}
\psfrag{A1}[l][l]{{\flabelsize A$_1$}}
\psfrag{A2}[Tl][Bl]{{\flabelsize A$_2$}}
\par\noindent\hspace{2em}
\begin{minipage}{.8\linewidth}
  \footnotesize
  \includegraphics[width=\linewidth,height=.25\textheight]{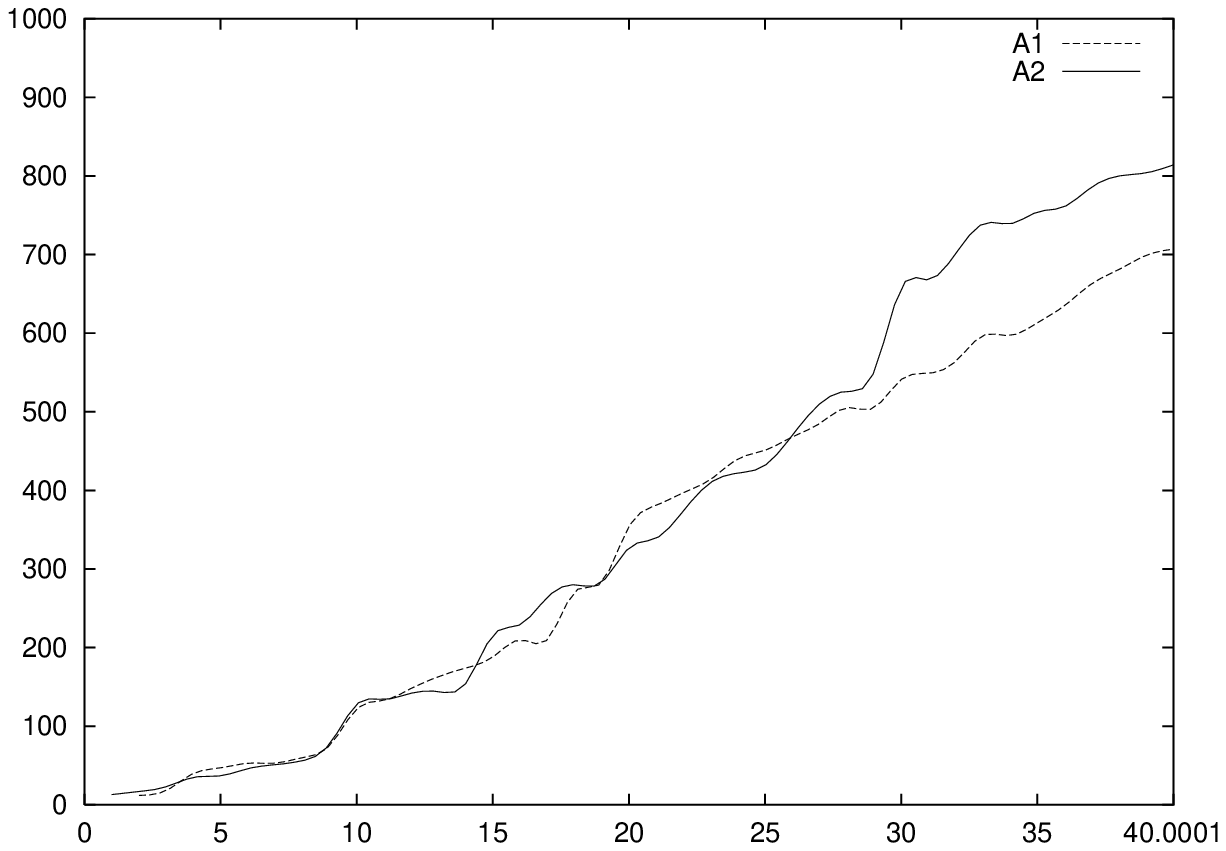}
\par\centering{\footnotesize (\ref{fig:results}a) Number of states \textit{S}}
\end{minipage}
\par\noindent\hspace{2em}
\begin{minipage}{.8\linewidth}
  \footnotesize
  \includegraphics[width=\linewidth,height=.25\textheight]{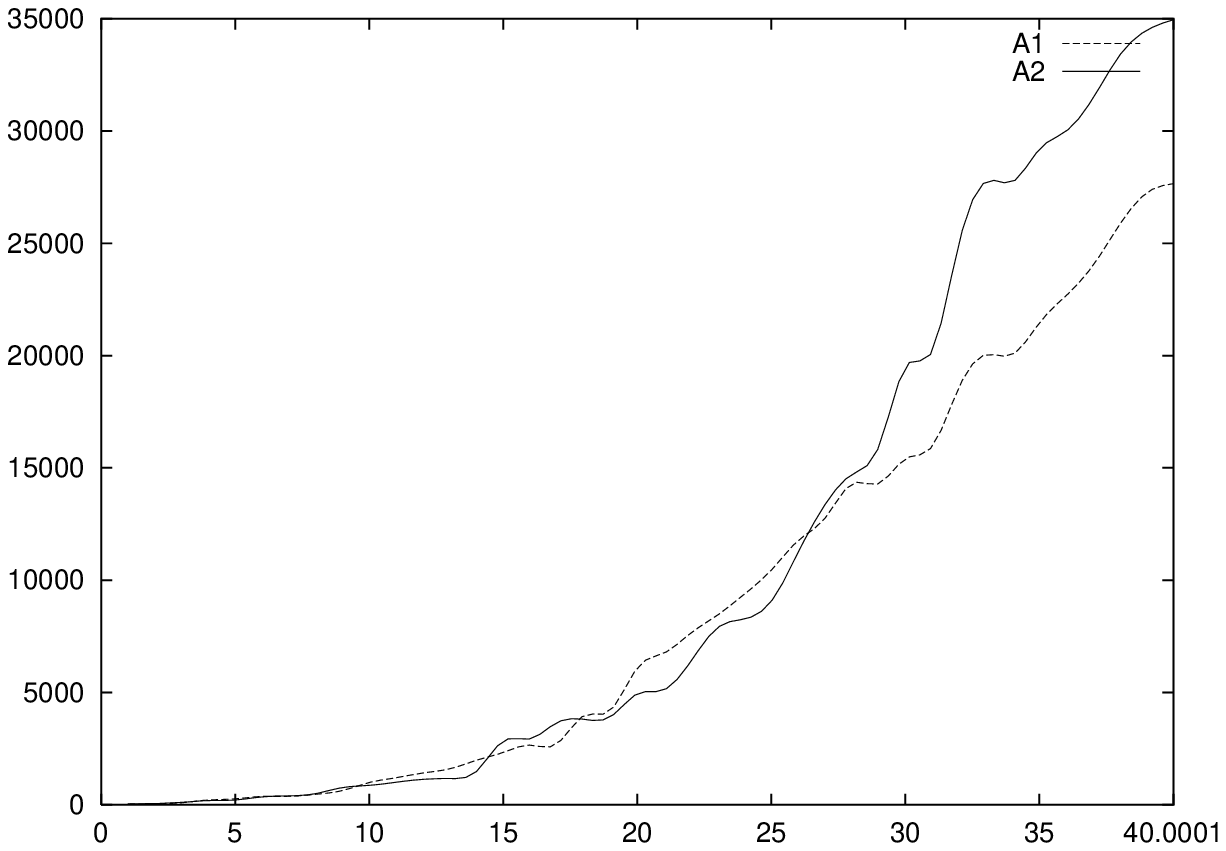}
\par\centering{\footnotesize (\ref{fig:results}b) Number of transitions \textit{Tr}}
\end{minipage}
\par\noindent\hspace{2em}
\begin{minipage}{.8\linewidth}
  \footnotesize
  \includegraphics[width=\linewidth,height=.25\textheight]{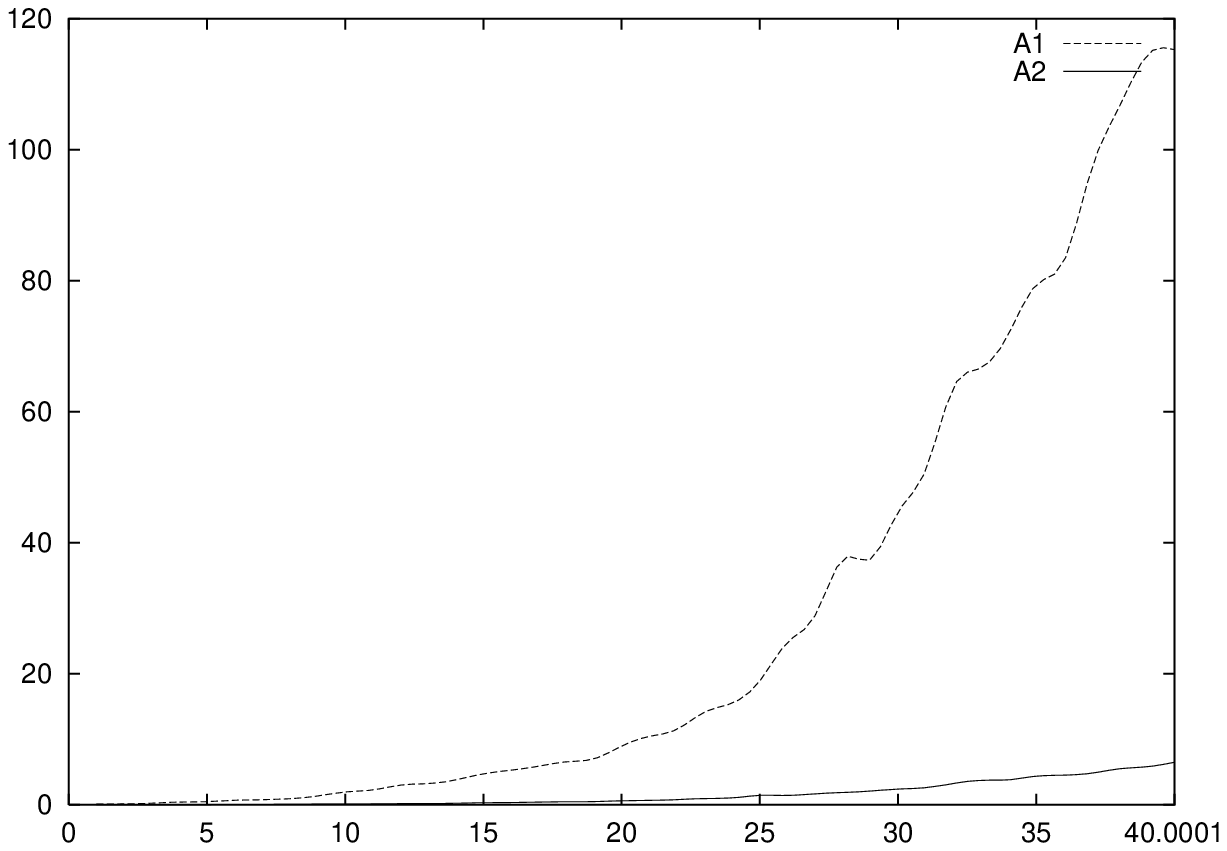}
\par\centering{\footnotesize (\ref{fig:results}c) Runtime in seconds \textit{t}}
\end{minipage}
\caption{%
	Comparison of two compilation algorithms.  A$_1$ is an
	transducer-based construction method \cite{Skut:ea:04}, A$_2$ the
	approach proposed in this paper.  The figures show
	the numbers of states (\ref{fig:results}a), the number of
	transitions (\ref{fig:results}b) and the runtime
	(\ref{fig:results}c) depending on the size of the input rule
	file ($n$ is the number of rules).  }
\label{fig:results}
\end{figure}

\section{Performance Evaluation}
\label{sec:perf}

To evaluate the performance of the new compilation method,
we measured the compilation time and the size of the resulting
structures for a set of homograph disambiguation rules in the format
described in section~\ref{sec:applications}. The results were compared
to the results achieved using a compiler that converts each rule into
an FST and then composes the FSTs and determinises
the transducer created by composition \cite{Skut:ea:04}. Both
algorithms were implemented in
\Cpp\ using the same library of FST/FSA classes, so the results solely
reflect the difference between the algorithms.

The figures (\ref{fig:results}a)--(\ref{fig:results}c) on
page~\pageref{fig:results} show the results of running both
implementations on a Pentium~4 1.7 GHz processor for rule sets of
different sizes. Figure (\ref{fig:results}a) shows the number of
states, (\ref{fig:results}b) the number of transitions, and
(\ref{fig:results}c) the compilation time. The numbers of states and
transitions for A$_2$, the bimachine-based approach proposed in this
paper, are the sums of the states and transitions, respectively, for
the left-to-right and right-to-left acceptors. The left-to-right FSA
typically has a much smaller number of states and transitions than the
right-to-left FSA (only 10\% of its states and 2--5\% of its
transitions) since it does not contain the regular expression for the
rule focus.

While the figures show a substantial decrease in runtime for the
bimachine construction method (A$_2$) compared to the FST-based
approach A$_1$ (only 6.48 seconds instead of 115.29 seconds for the
largest set of 40 rules in (\ref{fig:results}c)), the numbers of
states and transitions are slightly larger for the
bimachine. Typically the FSAs have about 25\% more states and 35\%
more transitions than the corresponding STs in our test set.  However,
an FSA takes up less memory than an FST as there are no emissions
associated with transitions and the output function $h$ can be encoded
in a very space-efficient way. As a result, the size of the compiled
structure in RAM was down by almost 30\% compared to the size of the
original transducer.

%% file: conclusions.tex
\section{Conclusion}

The rule compiler described in this paper presents an attractive
alternative to compilation methods that use FST composition and
complement in order to convert rewrite rules into finite-state
transducers.  The direct combination of context patterns into an
acceptor with final outputs makes it possible to avoid the use of
relatively costly FST algorithms. In the present implementation, the
only potentially expensive routine is the creation of the
deterministic acceptors for the context patterns. However, if the task
is to create a deterministic device, determinisation (in its more
expensive version for FSTs) is also required in the FST-based
approaches \cite{Skut:ea:04}. The experimental results presented in
section~\ref{sec:perf} show that compilation speed is not a problem in
practice. Should it become an issue, there is still room for
optimisation. The potential bottleneck due to DFSA determinisation can
be eliminated if we use a generalisation of the Aho-Corasick string
matching algorithm \cite{Aho:Corasick:75} in order to construct the
deterministic acceptor for the language $\Sigma^*\bigcup_{j\in I}
\beta_j \$_j$ while creating the  $\SimultMatch$ construct
\cite{Mohri:97a}.

By constructing a single deterministic device, we pursue a 
strategy similar to the compilation algorithms described by
Laporte~\shortcite{Laporte:97} and
Hetherington~\shortcite{Hetherington:01}. Our method shares some of
their properties such as the restriction of the rule focus $\phi$ to
one input symbol.
\footnote{As pointed out in section~\ref{sec:formalism}, this 
restriction does not pose a problem as the compiler is primarily
designed for rule-based tagging.}  However, it is more powerful as it
allows unrestricted (also cyclic) regular expressions in both the left
and the right rule context. The practical significance of this extra
feature is substantial: unlike phonological rewrite rules (the topic
of both Laporte and Hetherington's work), homograph disambiguation
does involve inspecting non-local contexts, which often pose a
difficulty to the 3-gram HMM tagger used to assign POS tags in our
system.

Although the use of our compiler is currently restricted to
hand-written rules, the extensions sketched in
section~\ref{sec:altercontrol} make it possible to use it in a machine
learning scenario (for both training and run-time application).

Our rule compiler has been applied successfully to a range of tasks in
the domain of speech synthesis, including homograph resolution,
post-lexical processing and phrase break prediction. In all these
applications, it has proved to be a useful and reliable tool for the
development of large rule systems.